\documentclass{article}

\PassOptionsToPackage{numbers, compress}{natbib}

\usepackage[preprint, nonatbib]{neurips_2026}

\usepackage[most]{tcolorbox}

\usepackage{amsmath}
\usepackage{listings}
\usepackage{xcolor}
\usepackage{comment}
\usepackage{url}
\usepackage{subcaption}
\usepackage{diagbox}

\usepackage{pgfplots}

\usepackage{booktabs}
\usepackage{adjustbox}
\usepackage[colorlinks=true,linkcolor=black,citecolor=black,urlcolor=blue,filecolor=blue]{hyperref}
\usepackage{graphicx} 
\usepackage{tikz}
\usetikzlibrary{positioning, calc, decorations.pathreplacing, arrows.meta, external}

\usepackage[backend=biber, maxbibnames=10, natbib=true]{biblatex}
\newtcbtheorem{hypothesis}{Hypothesis}{
  colback=blue!5!white,
  colframe=blue,
  arc=5pt,
  drop shadow,
  boxrule=1pt,
  fonttitle=\bfseries
}{hyp}

\title{Decoupling the Benefits of Subword Tokenization for Language Model Training via Byte-level Simulation}

\author{%
  Théo~Gigant\\
  Nous~Research \\
  \texttt{theo@nousresearch.com} \\
  \And
  Bowen~Peng \\
  Nous~Research \\
  \texttt{bloc@nousresearch.com} \\
  \And
  Jeffrey~Quesnelle \\
  Nous~Research \\
  \texttt{emozilla@nousresearch.com} \\
}

\begin{document}

\maketitle

\begin{abstract}

Subword tokenization is an essential part of modern large language models (LLMs), yet its specific contributions to training efficiency and model performance remain poorly understood.
In this work, we decouple the effects of subword tokenization by isolating them within a controlled byte-level pretraining pipeline.
We formulate and test hypotheses across various dimensions, including sample throughput, vocabulary scaling, and the linguistic prior of subword boundaries.
By simulating these effects in a byte-level setting, we refine our understanding of why subword models outperform raw byte models and offer insights to improve the pretraining of future byte-level and subword models.
Specifically, our experiments highlight the critical role of increased training throughput and the integration of subword boundaries as either explicit priors or inductive biases.

\end{abstract}

\section{Introduction}

Tokenization is an essential step of the Natural Language Processing pipeline, segmenting text into atomic units to be processed by language models.
Although state-of-the-art Large Language Models (LLMs) rely almost exclusively on subword algorithms like BPE or Unigram~\citep{sennrich_neural_2016, kudo_subword_2018}, there is no consensus on which specific properties of subword models enable this performance advantage~\citep{galle_investigating_2019, schmidt_tokenization_2024}.

Subword tokenization simultaneously dictates the allocation of compute to parts of the input sequence and the scaling of the model's vocabulary parameters by balancing vocabulary size, sequence length, and information density per token through the granularity of the tokens, or \emph{fertility} of the tokenizer.
Empirical evidence suggests that a larger vocabulary results on average in better downstream performances~\citep{tao_scaling_2024, huang_over-tokenized_2025} in part because it reduces the Kolmogorov complexity of tokenized sequences~\citep{chung_exploiting_2025}.
Subword tokens are also often viewed as a proxy for linguistic ``information units''~\citep{bostrom_byte_2020}.

Despite their prevalence, recent literature has highlighted significant issues stemming from subword tokenizers, including ``character-blindness''~\citep{ciaccio_beyond_2025, cosma_strawberry_2025}, language-dependent performance disparities~\citep{rust_how_2021}, inadequacies with prefix forms~\citep{lerner_unlike_2025}, tokenization ambiguity~\citep{kudo_subword_2018, provilkov_bpe-dropout_2020}, and weaknesses linked to under-trained tokens~\citep{land_fishing_2024}.

Character, or byte-level language models~\citep{clark_canine_2022, minixhofer_bolmo_2025} have been proposed as an alternative to subword language models, in part to address these issues.
Sometimes wrongly described as \emph{tokenizer-free}, these models usually rely on characters as defined by the Unicode standard~\citep{unicode_consortium_unicode_1991}, or bytes resulting from the UTF-8~\citep{yergeau_utf-8_2003} encoding of text.
While solving some of the aforementioned subword-related problems, these byte-level language models consistently struggle to match the training efficiency and downstream performance of their subword-based counterparts.
This performance gap between byte-level and subword models is typically attributed to some ``benefits'' of subword tokenization, which are typically analyzed in aggregate.
To the best of our knowledge, there have been no successful attempts to isolate and quantify their decoupled contributions.
For example, a larger vocabulary not only increases embedding capacity, but also reduces sequence length, thereby increasing the effective sample throughput during training.
Furthermore, subword boundaries may provide a structural prior that aligns with human semantics, aiding generalization in ways that raw bytes do not.

In this paper, we suggest hypotheses as to what effects subword tokenization methods have on training dynamics, and we conduct a set of experiments to try to isolate and quantify them by artificially reproducing these effects for training byte-level language models.

\section{Preliminaries}

\subsection{Subword tokenization}

Byte-Pair Encoding (BPE)~\citep{sennrich_neural_2016} is a bottom-up subword tokenization method based on the BPE grammar-based compression algorithm~\citep{gage_new_1994}.
It is the \textit{de facto} standard tokenization method used with LLMs.
It comes as the default tokenization method in the most popular LLM training frameworks~\citep{shoeybi_megatron-lm_2020, liang_torchtitan_2025, axolotl_maintainers_and_contributors_axolotl_2023, daniel_han_unsloth_2023}, due to highly optimized implementations\footnote{Such as \url{https://github.com/huggingface/tokenizers} or \url{https://github.com/openai/tiktoken}}.
Its dominance can also be attributed to the legacy of open-source LLMs that had a great impact on industry and academia, such as GPT-2~\citep{radford_language_2019}, LLaMA~\citep{touvron_llama_2023} and Mistral~\citep{jiang_mistral_2023}.

A popular alternative is unigram tokenization~\citep{kudo_subword_2018}, a top-down subword tokenization method based on a unigram language model, which creates tokens that align better with morphology~\citep{bostrom_byte_2020} and allows subword regularization~\citep{kudo_subword_2018}.
This method is more rarely encountered in practice, due to the more costly and difficult implementation.

\subsection{Byte-level language models}

Contrary to LLMs using static subword tokenization, byte-level LLMs have a more fine-grained access to single bytes of the input.
These models usually involve a method to compress or downsample the byte sequences to align the FLOPs-per-input-byte cost with subword models.
These include, for instance, static downsampling with strided convolutions~\citep{clark_canine_2022}, or dynamic downsampling using lightweight local encoders~\citep{pagnoni_byte_2025, hwang_dynamic_2025, minixhofer_bolmo_2025}.

In contrast with these works, in this work we do not use downsampling in the architecture and process UTF-8-tokenized sequences with a standard architecture for subword-tokenized sequences, namely the LLaMA-3 architecture~\citep{grattafiori_llama_2024}.

\section{Related Works}

Previous works have studied the effects of subword tokenization for language model training.
\citet{galle_investigating_2019} and~\citet{zouhar_formal_2023} empirically showed that a BPE tokenizer with a higher compression ratio results in higher downstream performance on machine translation tasks.
\citet{chung_exploiting_2025} quantified the complexity of tokenized text via an estimate of the Kolmogorov complexity, showing that increasing the vocabulary size of a BPE tokenizer increases performance as a consequence of a reduction in the complexity of tokenized sequences.
\citet{schmidt_tokenization_2024} developed a tokenization scheme that compresses sequences more than BPE, while resulting in worse downstream performance, challenging the idea that the effectiveness of BPE comes only from its compression effect.

In this paper, we formulate and test hypotheses covering various aspects of subword tokenization, including computational efficiency, structural inductive biases and changes to the optimization objective.

\section{Hypotheses}

\label{sec:hyp}

We formalize the potential drivers of the subword-byte performance gap into the following testable hypotheses, categorized by their effects on model training and representation.

\subsection{Computational and Scaling Efficiency}

The advantages most commonly attributed to tokenization relate to sequence compression.
By reducing sequence lengths and expanding the vocabulary, tokenization fundamentally alters the structural dimensionality of the model's input and the marginal computational cost per bit of processed data.

\begin{hypothesis}{}{flops-free-params}
    {Increasing subword vocabulary size improves performance by scaling model capacity (via the embedding layer) at a marginal increase in FLOPs.}
\end{hypothesis}

Token embeddings are usually implemented as look-up tables, accessed in constant time.
As noted by~\citet{tao_scaling_2024}, large vocabularies improve model performance, and most of the computational overhead of adding vocabulary parameters is related to the output layer.

\begin{hypothesis}{}{sample-throughput}
    {At an isoFLOP budget, subword models process significantly more raw information (in bits) per gradient step than byte-level models, leading to superior sample throughput and better downstream performance.}
\end{hypothesis}

\subsection{Structural Inductive Biases}

Subword tokenization injects ``human-centric'' structure into the sequence before the model ever sees it.
We hypothesize that this acts as a powerful prior and could be leveraged as an inductive bias to improve training.

\begin{hypothesis}{}{subword-bounds-prior}
    {Subword end boundaries provide a non-causal prior that simplifies the modeling task by leaking future bytes.}
\end{hypothesis}

Unlike UTF-8 tokenization, which is strictly causal, subword tokenizers require a ``look-ahead'' to determine optimal boundaries~\citep{minixhofer_bolmo_2025}.
This effectively provides the model with a ``hint'' about the future byte distribution, creating an inherently easier prediction task.

\begin{hypothesis}{}{subword-bounds-bias}
    {Subword boundaries act as inductive biases that align the model with semantically meaningful units.}
\end{hypothesis}

\begin{hypothesis}{}{subword-dist-prior}
    {Subword distances act as inductive biases that align the model’s attention mechanism and positional encodings with semantically meaningful units.}
\end{hypothesis}

In subword LLMs, positional encodings represent distances between subwords; in byte-level models, they usually represent character distances, which may lack direct semantic utility.

\subsection{Optimization Objective}

Finally, we consider how the choice of tokenization shifts the nature of the prediction task itself.

\begin{hypothesis}{}{cross-entropy-per-subword}
    {Minimizing cross-entropy per subword is a more effective proxy for language modeling than minimizing cross-entropy per byte.}
\end{hypothesis}

\begin{hypothesis}{}{next-subword-prediction}
    {Subword prediction functions as a form of variable-length multi-token prediction, which has been shown to improve representation learning compared to single-token prediction.}
\end{hypothesis}

Predicting a single subword is equivalent to predicting a byte $n$-gram at once.
This aligns with recent findings that multi-token prediction heads can improve downstream performance~\citep{gloeckle_better_2024}.

\section{Methodology}
\label{sec:exp}

We propose experiments intended to replicate one by one the effects induced by subword tokenization linked to the hypotheses we suggested.
These effects are added to a 1.7B parameters byte-level language model pretraining pipeline, which will be compared to a baseline byte-level language model.

In the following experiments, most hyperparameters remain unchanged.
All changes made on the input and output, or on the architecture of the model, are designed to introduce negligible computational overhead.
We are using a standard LLaMA-3 architecture~\citep{grattafiori_llama_2024} trained with the TorchTitan framework~\citep{liang_torchtitan_2025}.
Models are trained on the fineweb-edu dataset~\citep{penedo_fineweb_2024} tokenized into UTF-8 bytes.
Sequences are also tokenized with the LLaMA-3 BPE tokenizer to provide byte-level subword boundaries.
All comparisons between models are done using the same bits-per-byte cross-entropy loss, computed on a separate validation subset of fineweb-edu.
Hyperparameters are detailed in the Appendix \ref{apx:hp}.

\subsection{Scaling vocabulary parameters}

To test Hypothesis \ref{hyp:flops-free-params}, we introduce multi-head $n$-gram embedding tables to simulate the larger input vocabulary of a subword LLM. 
This method is similar to recent $n$-gram embedding methods~\citep{huang_over-tokenized_2025, liu_scaling_2026, cheng_conditional_2026}, but we introduce them only in the input layer.
Our implementation is derived from the \emph{engram} demo implementation \footnote{\url{https://raw.githubusercontent.com/deepseek-ai/Engram/refs/heads/main/engram_demo_v1.py}}.

Hyperparameters are chosen to introduce around $70$M additional parameters to the byte-level LLM, matching the embedding table of a subword LLM using the same architecture with a vocabulary of $35$k tokens.

\begin{figure}[h]
    \centering
    \resizebox{0.66\linewidth}{!}{\input{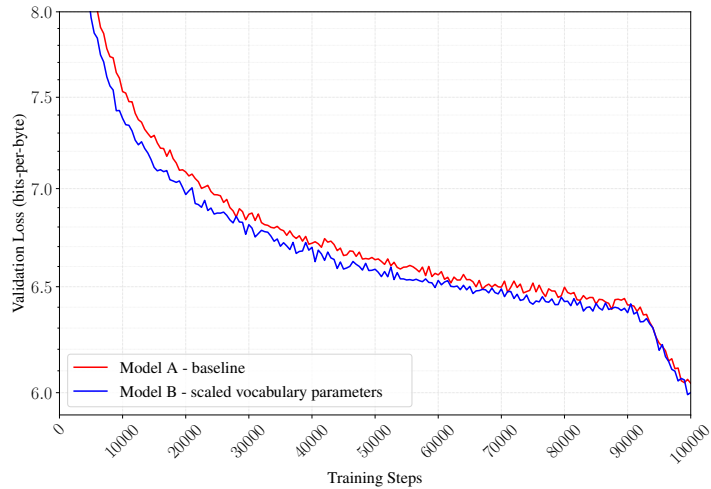}}
    \caption{Validation loss when scaling input embedding parameters}
    \label{fig:engram}
\end{figure}

Figure \ref{fig:engram} highlights the small increase in performance associated with scaling input embedding parameters.
While these results suggest that Hypothesis \ref{hyp:flops-free-params} does not explain the significant performance gap between subword and byte-level language models, scaling vocabulary-like parameters remains a promising direction to improve language models, as exemplified in recent literature~\citep{roy_n-grammer_2022, huang_over-tokenized_2025, cheng_conditional_2026, liu_scaling_2026}.

\subsection{Artificially increasing the training sample throughput}

Subword tokenization results on average in around $4$ times fewer tokens compared to UTF-8 tokenization\footnote{We measure an average of $4.75$ bytes-per-token on $50,000$ samples from fineweb-edu tokenized with the LLaMA-3 tokenizer.}.
At isoFLOPs, the sample throughput during training is $4$ times higher using the same architecture.
To simulate this behavior, we compress the sequences by a factor of $4$ to train a byte-level LLM at the same isoFLOPs sample throughput as a subword LLM.

Given a sequence of length $4\cdot L$, we segment it into contiguous chunks of $4$ bytes, resulting in a sequence of shape ($L, 4$).
In the input layer, the model sums the embeddings of the $4$ contiguous bytes in each chunk.
In all hidden layers, the behavior is unchanged, and the model is effectively processing a sequence of $L$ latent tokens, containing information from $4\cdot L$ input tokens.
The model output has a shape ($L, V$) with $V$ the size of the vocabulary.
The loss is computed as the cross-entropy between this prediction and the first byte of the next chunk, \textit{i.e.} next-byte prediction.

After $50$k steps using this method to artificially increase sample throughput by $4$ times, we continue pretraining this model with the baseline regime, using sequences of length $L$.
During the first $50$k steps, the baseline model $A$ sees on average $4$ times less samples compared to model $B$, but the same number of tokens, effectively simulating the larger sample throughput of subword language models.
After $50$k steps, both models are trained under the same conditions.

\begin{figure}[h]
    \centering
    \resizebox{0.66\linewidth}{!}{\input{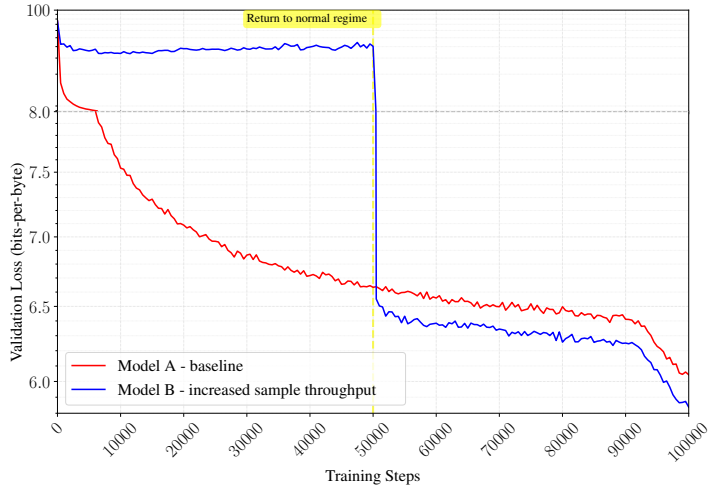}}
    \caption{Validation loss when scaling sample throughput by $4$ times for $50$k steps}
    \label{fig:samplethroughput}
\end{figure}

Figure \ref{fig:samplethroughput} illustrates a significant gain resulting from the increase in sample throughput, even if performed for only $50$k steps.
Rapidly after falling back to the normal regime, model $B$ crosses the performance of the baseline model $A$, and soon stabilizes at the same slope.
This experiment strongly supports Hypothesis \ref{hyp:sample-throughput}.

\subsection{Giving subword boundaries as a prior}

Subword tokenization segments the input text into contiguous chunks based on frequencies of $n$-grams in the training corpus.
This process requires access to the full sequence and thus leaks future information into past tokens~\citep{minixhofer_bolmo_2025}.
A subword LLM is optimized for next-token prediction given a correctly segmented input.
On the other hand, byte-level LLMs are usually strictly causal.
We posit that having access to the subword segmentation boundaries makes the prediction task easier.
For example, by design of pre-tokenization, whitespace characters are always following an end-of-subword boundary.
On the other hand, the start-of-subword boundaries do not leak future bytes information, but can provide the model with structural prior.

In the following experiment, models $B$ and $C$ have access to a binary sequence of start-of-subword and end-of-subword boundaries, respectively, whose embeddings are added to the input byte embeddings.

\begin{figure*}[ht]
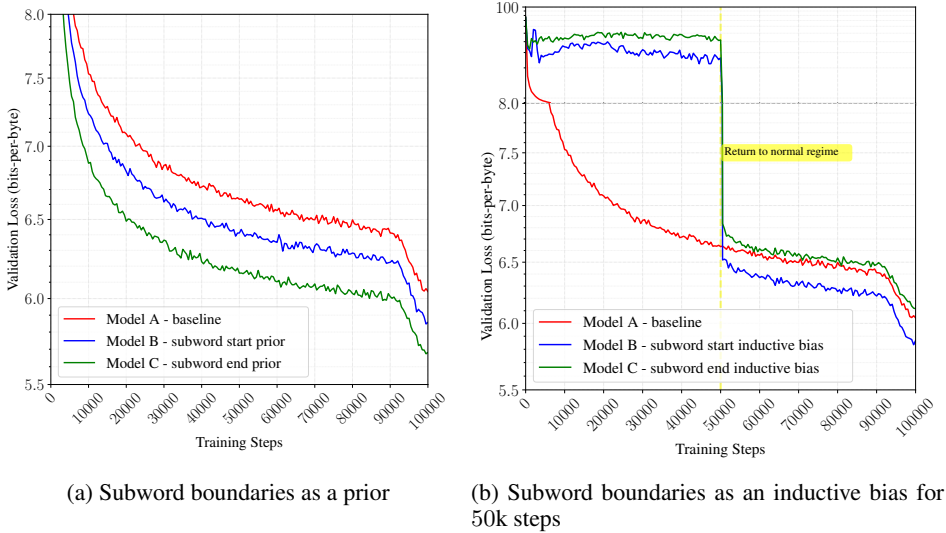

    \centering
    \begin{subfigure}[t]{0.45\textwidth}
        \centering
        \resizebox{\textwidth}{!}{\input{figures/subword-boundaries.pgf}}
        \caption{Subword boundaries as a prior}
        \label{fig:subword-boundaries}
    \end{subfigure}
    \begin{subfigure}[t]{0.45\textwidth}
        \centering
        \resizebox{\textwidth}{!}{\input{figures/subword-boundaries-prior.pgf}}
        \caption{Subword boundaries as an inductive bias for $50$k steps}
        \label{fig:subword-boundaries-prior}
    \end{subfigure}
    \caption{Validation loss when providing the start or end of subword boundaries}
    \label{fig:subword-boundaries-figures}
\end{figure*}

Figure \ref{fig:subword-boundaries} shows the significant performance boost resulting from the access to the subword segmentation boundaries, supporting Hypothesis \ref{hyp:subword-bounds-prior}.
Specifically, end-of-subword boundaries offer a larger advantage compared to start-of-subword boundaries, as they leak future information.
Start-of-subword boundaries also improve the performance of the model, suggesting that the statistical prior they provide is a useful inductive bias for the model.
In order to test that hypothesis, we train the models with access to the subword boundaries only at training-time, and remove the boundary information at validation-time.
After $50$k steps, we also remove the access to the subword boundaries for training and resume pretraining following the baseline regime.

While subword end boundaries are more useful as a prior than subword start boundaries (\textit{c.f.} Figure \ref{fig:subword-boundaries}), they do not provide a useful inductive bias in this setting as evidenced by Figure \ref{fig:subword-boundaries-prior}, probably because the model is relying too much on this prior.
On the other hand, subword start boundaries do not leak future information, and provide a prior that improves the model performance in this setting.
These observations support Hypotheses \ref{hyp:subword-bounds-prior} and \ref{hyp:subword-bounds-bias}.

\subsection{Giving subword distances as a prior}

Similarly,~\citet{gelberg_extending_2025} showed that RoPE positional encoding acts as a prior that can be removed later during training.
In subword LLMs, the positional encoding is using subword distances, when byte-level LLMs use byte distances.
To simulate the position prior of the subword positions in the latter, we replace the byte position encoding with subword position encoding in model $B$.
Subsequent bytes that are part of the same subword use the same repeated position.
This setting does not leak future byte information, as it is effectively using the subword start boundaries information.

\begin{figure*}[ht]
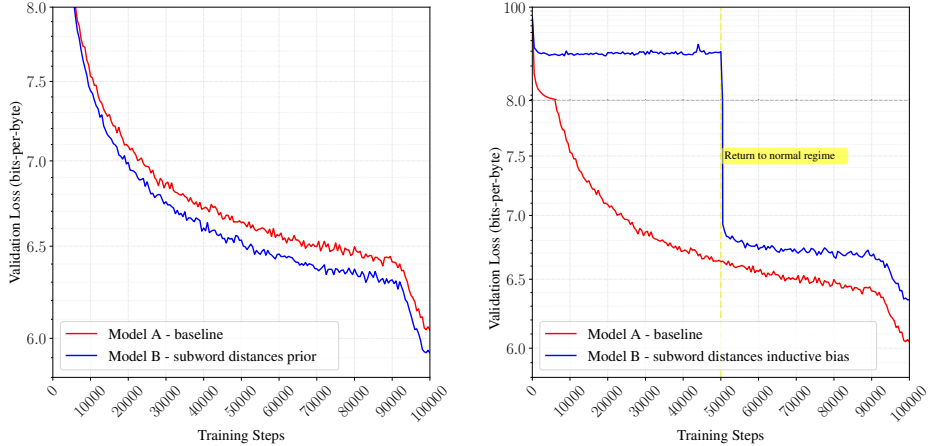

    \centering
    \begin{subfigure}[t]{0.45\textwidth}
        \centering
        \resizebox{\textwidth}{!}{\input{figures/subword-distances.pgf}}
        \caption{Subword distances as a prior}
        \label{fig:subword-distances}
    \end{subfigure}
    \begin{subfigure}[t]{0.45\textwidth}
        \centering
        \resizebox{\textwidth}{!}{\input{figures/subword-distances-prior.pgf}}
        \caption{Subword distances as an inductive bias for $50$k steps for training only}
        \label{fig:subword-distances-prior}
    \end{subfigure}
    \caption{Validation loss when using subword distances in the positional embedding}
    \label{fig:subword-distances-figures}
\end{figure*}

We perform another experiment in which this prior is given only during training and removed after $50$k steps, returning to the baseline training regime afterwards.

Figures \ref{fig:subword-distances} and \ref{fig:subword-distances-prior} suggest that subword distances can be a useful prior, but do not constitute a strong inductive bias in this setting.
Considering the previous section, we conclude that subword boundaries constitute stronger prior, and inductive biases, than subword distances, highlighting the lesser relative significance of Hypothesis \ref{hyp:subword-dist-prior} compared to the previous Hypotheses.

\subsection{Optimizing cross-entropy per subword}

The cross-entropy loss for predicting a sequence of subwords $(t_m)_{m\leq M}$ with a model $\theta$ is defined as
$$CE_\text{subword}(\theta, (t_m)_{m\leq M})=-\frac{1}{M}\sum_{m\leq M}\log(P_\theta(t_m|(t_k)_{k<m}))$$
With the same sequence, but tokenized as UTF-8 bytes $(b_n)_{n\leq N}$, the default cross-entropy becomes
$$CE_\texttt{UTF-8}(\theta, (b_n)_{n\leq N})=-\frac{1}{N}\sum_{n\leq N}\log(P_\theta(b_n|(b_k)_{k<n}))$$

However, by decomposing a subword $t$ into the $k$ bytes it contains $(b_i)_{i\leq k}$, we have $P_\theta(t) = \Pi_{i\leq k}(P_\theta(b_i|(b_j)_{j< i}))$

Thus,
$$CE_\texttt{UTF-8}(\theta, (b_n)_{n\leq N})=-\frac{1}{N}\sum_{m\leq M}\log(P_\theta(t_m|(t_k)_{k<m})) = \frac{M}{N} \cdot CE_\text{subword}(\theta)$$

Instead of optimizing for the best cross-entropy per subword, the baseline target for byte-level LLM optimizes for cross-entropy per byte, scaling the loss by a dynamic factor $\frac{M}{N}<1$.
In order to see if this difference has any consequence on training, we use the cross-entropy per subword as a target to train a byte-level LLM and compare to the baseline.

\begin{figure}[h!]
    \centering
    \resizebox{0.66\linewidth}{!}{\input{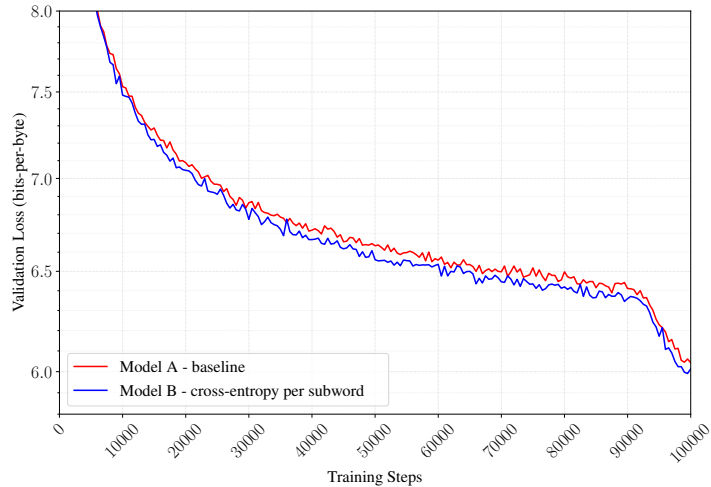}}
    \caption{Validation loss when optimizing for cross-entropy per subword}
    \label{fig:loss-per-subword}
\end{figure}

Figure \ref{fig:loss-per-subword} shows very little improvement compared to baseline, suggesting that Hypothesis \ref{hyp:cross-entropy-per-subword} has minimal effects at this scale.

\subsection{Optimizing next subword prediction}

Instead of predicting one byte at a time, a subword LLM predicts a subword, usually containing multiple bytes.
Arguably, this is analogous to multi-token prediction \cite{gloeckle_better_2024}, which was shown to improve pretraining of LLMs, especially for models with more than $1$ billion parameters.

We train byte-level model $B$ using a subword output vocabulary, optimizing for the cross-entropy computed using the next subwords, predicted from the end-of-subword bytes.
After $50$k steps, we return to the baseline pretraining regime.

\begin{figure}[h]
    \centering
    \resizebox{0.66\linewidth}{!}{\input{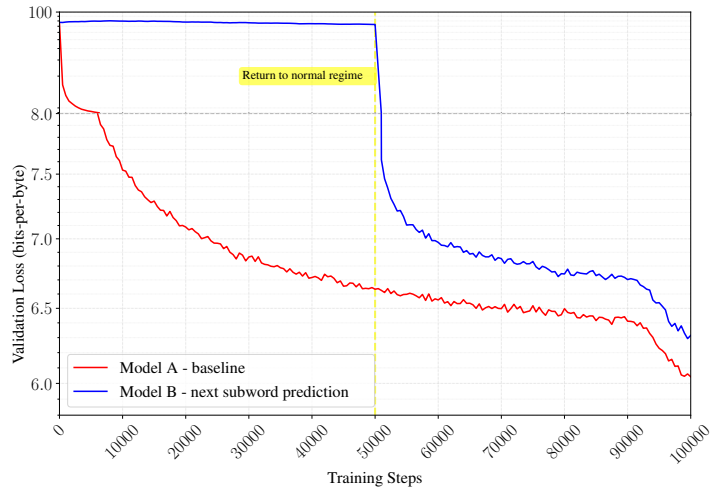}}
    \caption{Validation loss when optimizing for next subword prediction for $50$k steps}
    \label{fig:next-subword-prediction}
\end{figure}

Figure \ref{fig:next-subword-prediction} illustrates that the next subword prediction task is a worse objective to train a language model at this scale compared to next byte prediction, rejecting Hypothesis \ref{hyp:next-subword-prediction}.

\section{Summary}

The experiments we conducted suggest that the superior performance of subword language models compared to byte-level language models involve multiple effects at different magnitudes.
Specifically, the effects related to  Hypotheses \ref{hyp:sample-throughput}, \ref{hyp:subword-bounds-prior} and \ref{hyp:subword-bounds-bias} are the most noticeable at this scale.
By replicating these effects in isolation, we observe a significant improvement for pretraining byte-level language models.
Interestingly, these hypotheses are related to different aspects of subword tokenization.
The increased sample throughput (Hypothesis \ref{hyp:sample-throughput}) is a direct consequence of the compression capabilities of subword tokenization.
This is usually the biggest drawback that hinders the competitiveness of byte-level language models, such that state-of-the-art byte-level language models come with methods to compress the byte sequences and thus increase sample throughput closer to the subword counterpart~\citep{clark_canine_2022, pagnoni_byte_2025, hwang_dynamic_2025, minixhofer_bolmo_2025}.
On the other hand, prior knowledge of subword boundaries (Hypotheses \ref{hyp:subword-bounds-prior} and \ref{hyp:subword-bounds-bias}) has strong connections to subwords being good approximations of English semantic units.
As exemplified by~\citet{schmidt_tokenization_2024}, the compression aspect cannot completely explain the efficiency of subword LLMs.
Subwords created with Unigram, and BPE to a lesser extent, align well with morphological reference segmentations \cite{bostrom_byte_2020}, explaining why we empirically observed that they provide a useful inductive bias during byte-level language model pretraining.

Our tests to replicate the effects linked to Hypotheses \ref{hyp:flops-free-params}, \ref{hyp:subword-dist-prior}, \ref{hyp:cross-entropy-per-subword} and \ref{hyp:next-subword-prediction} either perform worse or do not show a significant change compared to the baseline, suggesting that these effects are not perceptible at this scale.
However, their significance could be different at different scales. For instance, we observed a larger gap for experiments linked with Hypothesis \ref{hyp:flops-free-params} for smaller models ($68$M parameters experiments are included in Appendix \ref{fig:small-xp}).

\section{Conclusion}

In this paper, we proposed hypotheses regarding the effects that subword tokenization is having on language modeling.
Through experiments simulating these effects in a pipeline for pretraining byte-level language models, we try to isolate these effects and quantify the improvement they provide.
In particular, we highlight the importance of increasing the training sample throughput, and giving the subword boundaries as a prior or as inductive biases.

We believe that a better understanding of these effects will prove useful to both improve subword tokenization and byte-level language model pretraining.
For example,~\citet{minixhofer_bolmo_2025} recently proposed a method to continue the pretraining of a subword LLM as a byte-level LLM, effectively taking advantage of the beneficial effects of subword tokenization during the first stage of byte-level language model pretraining.
\citet{zheng_proxy_2026} trained LLMs with inputs mixing raw unicode with sequences compressed using subword tokenization, neural compression or gzip, showing better isoFLOPs byte-level performance at scales exceeding 4B parameters, compared with baseline byte-level pretraining.
A better understanding of these effects in isolation could allow researchers to improve on some of them, for instance by using different tokenization schemes for different purposes, or even scale some of these effects, similarly to the recent works studying new scaling directions for vocabulary-like parameters decoupled from the model's vocabulary~\citep{huang_over-tokenized_2025, liu_scaling_2026, cheng_conditional_2026}.

\section{Limitations and Future Work}

\label{sec:lims}

While our controlled simulations provide valuable insights into the decoupled benefits of subword tokenization, this study has several limitations that present opportunities for future research.

To maintain computational feasibility while exploring a wide range of hypotheses, several of our key experimental interventions, such as artificially increasing sample throughput, injecting subword boundary priors, enforcing subword distance priors and optimizing for the next subword prediction objective, were introduced for only the first $50$k training steps before reverting to the baseline byte-level training regime.
While this setting was sufficient to observe significant shifts in validation loss and training dynamics in some settings, the behavior of these priors could be different at different model scales and intervention duration.
It remains an open question whether the performance gains, or the lack of them, observed in models exposed to these training interventions compound, plateau, or diminish when maintained throughout a complete, full-scale pretraining run.

A core methodological choice in this work was to replicate the effects induced by subword tokenization one by one.
By artificially isolating these variables, we successfully quantified their individual contributions to the subword-byte performance gap.
However, this decoupled approach does not account for the complex interplay between these mechanisms.
For instance, it is highly likely that the benefits of an increased training sample throughput and the structural inductive biases of subword boundaries  interact during standard subword language model training.
Future work should investigate these compounding effects to determine whether these isolated variables act additively, synergistically, or redundantly when combined into a single byte-level architecture.

Our experiments were conducted on a $1.7$B parameter language model trained exclusively on the English-centric fineweb-edu dataset tokenized into UTF-8 bytes.
As noted in our discussion, the significance of certain hypotheses may shift at larger, or smaller, parameter scales.
Furthermore, because English subwords naturally align well with morphological segmentations, the strength of the inductive biases provided by subword boundaries might differ substantially when modeling languages with different morphological structures~\citep{drzik_importance_2026}.
Expanding this byte-level simulation framework to highly multilingual datasets and larger model scales remains a promising direction for future research.

Finally, our work revolved around the perspective of reducing the gap between subword models and byte models, however, some of the insights could be leveraged to improve subword models further.

\printbibliography

\newpage
\appendix

\section{Hyperparameters}

\label{apx:hp}

\begin{table}[h]
    \centering
    \caption{Training hyperparameters}
    \begin{tabular}{l|c}
        Hyperparameter & Value \\
         \hline
         Architecture & LLaMA-3~\citep{grattafiori_llama_2024} \\
         \# parameters & $1,747,060,736$ \\
         Dimension & $2,048$ \\
         \# layers & $32$ \\
         \# heads & $32$ \\
         \# KV heads & $8$ \\
         Optimizer & AdamW \\
         $\beta_1, \beta_2$ & $0.9$, $0.95$ \\
         Weight decay & 0.1 \\
         Global batch size & $128$ \\
         Sequence length & $8,192$ \\
         Positional encoding & RoPE \\
         Learning rate scheduler & Warmup-Stable Cosine Decay \\
         Learning rate (peak) & $3*10^{-4}$ \\
         Training steps & $100,000$ steps \\
         LR warmup & $2,000$ steps \\
         LR decay & $10,000$ steps \\
         \hline
    \end{tabular}
    \label{tab:hyperparameters}
\end{table}

\begin{table}[h]
    \centering
    \caption{Multihead $n$-gram embedding hyperparameters}
    \begin{tabular}{l|c}
        Hyperparameter & Value \\
         \hline
         \# extra parameters & $71,065,088$ \\
         Maximum order $n$ of $n$-grams & $3$ \\
         Base $n$-gram vocabulary size at each order & $15,000$ \\
         Token vocabulary size & $512$ \\
         Embedding dimension & $2048$ \\
         \# heads & $8$ \\
         \hline
    \end{tabular}
    \label{tab:ngramparams}
\end{table}

Runs are performed on B200 GPUs for around 160 GPU-hours per run.

\section{Smaller-scale experiments}

\label{fig:small-xp}

\begin{figure*}[ht]
    \centering
    \begin{subfigure}[t]{0.45\textwidth}
        \centering
        \includegraphics[width=\linewidth]{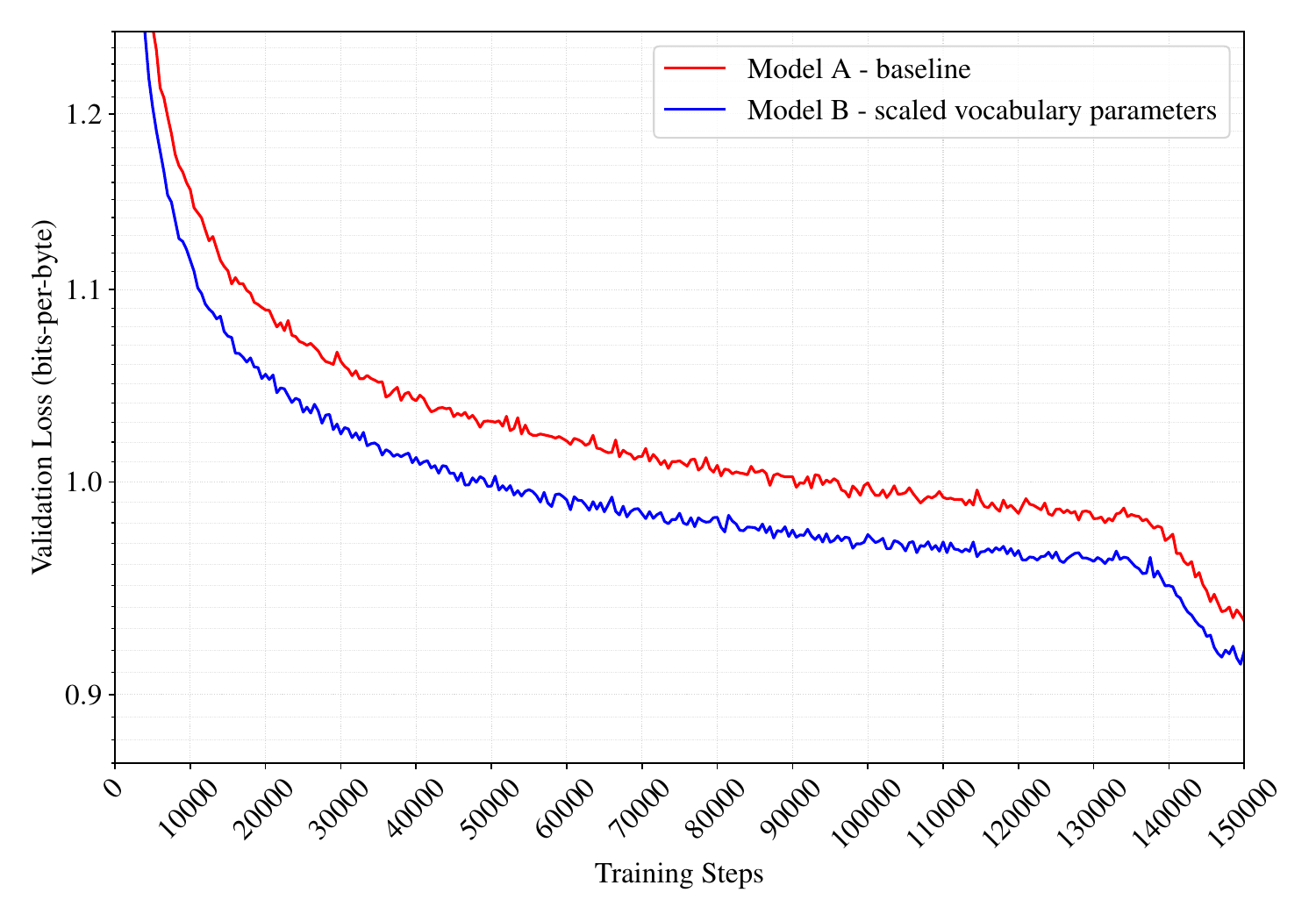}
    \caption{Scaling input embedding parameters}
    \label{fig:engram-small}
    \end{subfigure}
    \begin{subfigure}[t]{0.45\textwidth}
        \centering
        \includegraphics[width=\linewidth]{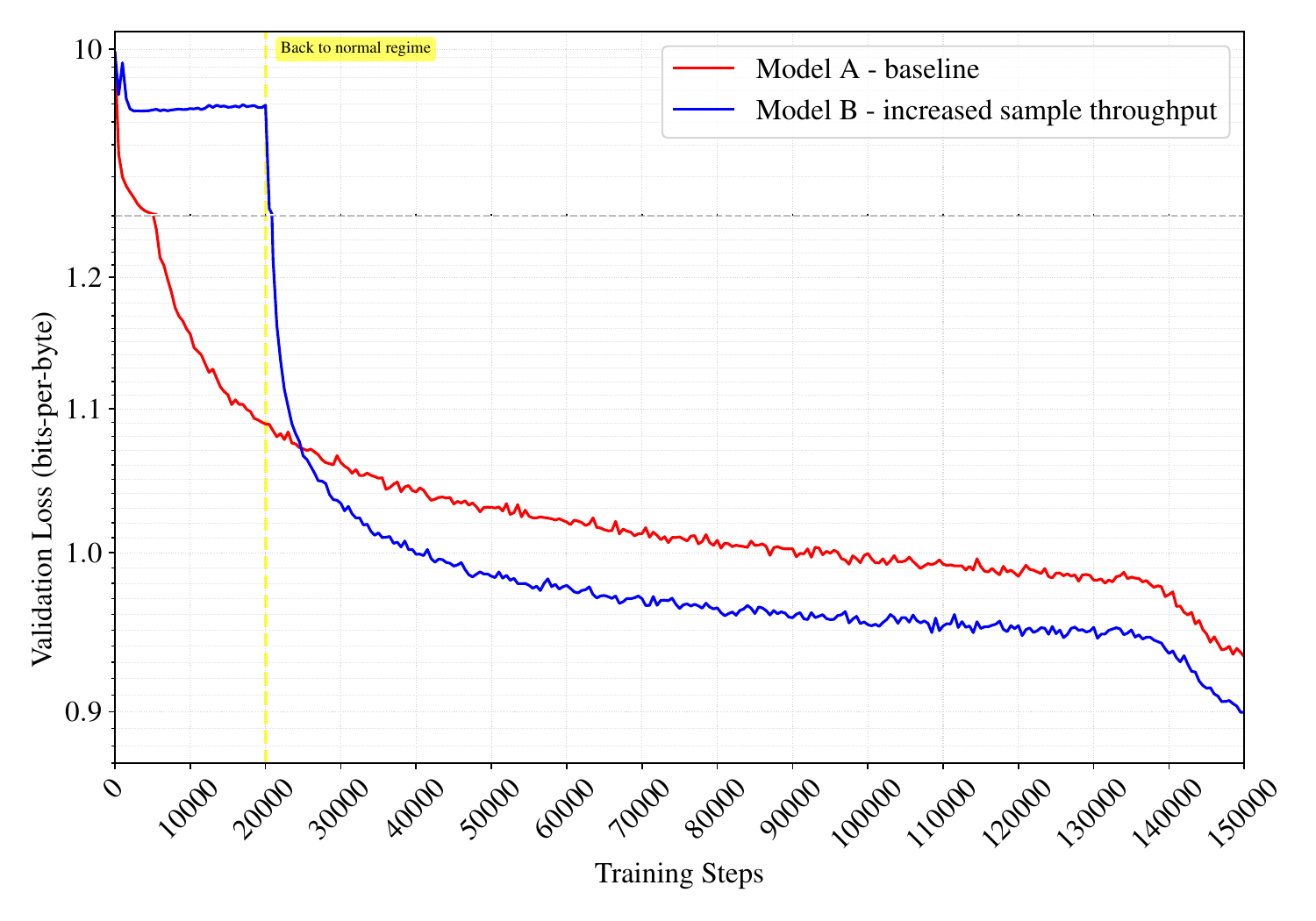}
    \caption{Scaling sample throughput by $4$ times for $20$k steps}
    \label{fig:samplethroughput-small}
    \end{subfigure}
    \begin{subfigure}[t]{0.45\textwidth}
        \centering
        \includegraphics[width=\linewidth]{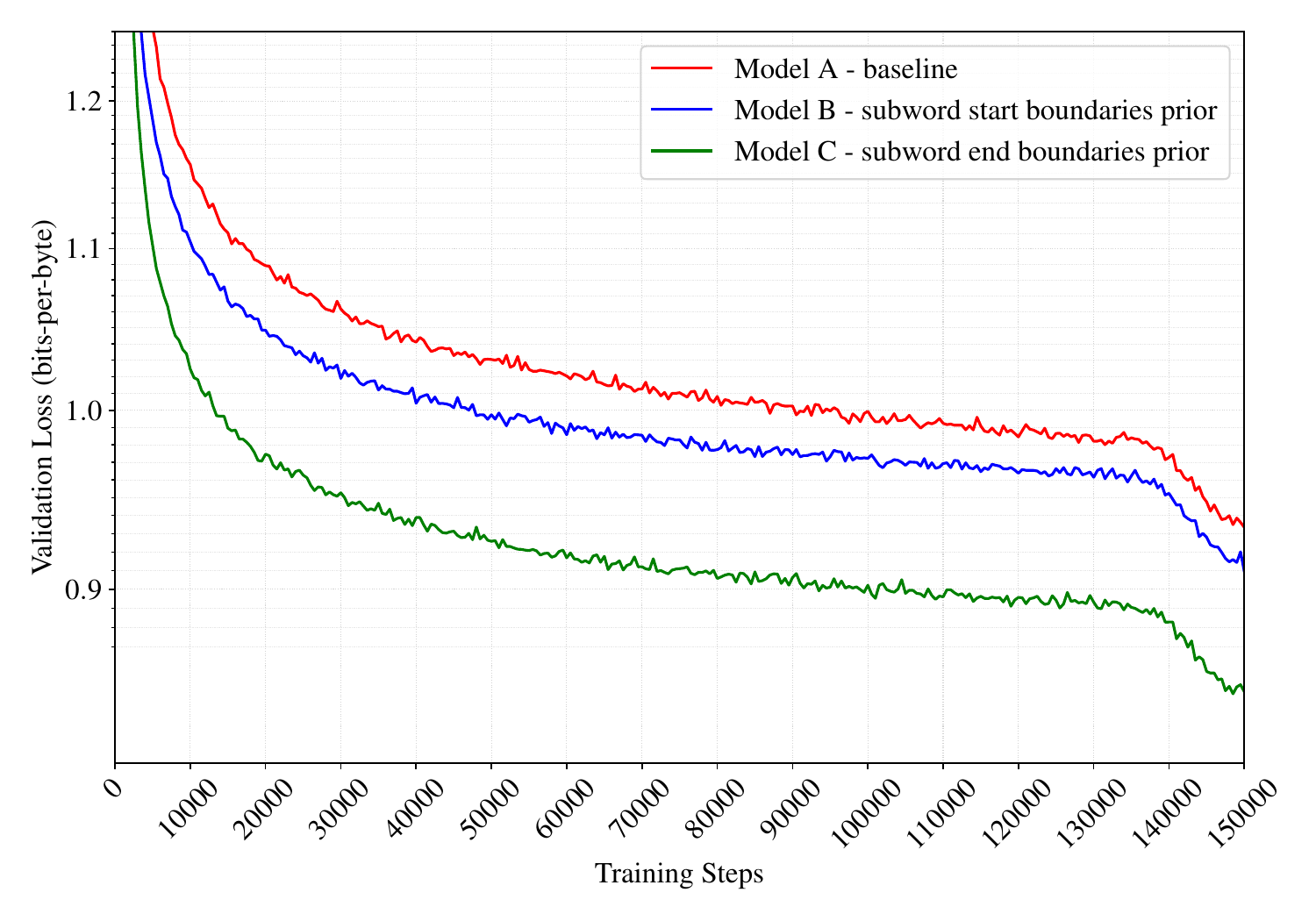}
        \caption{Subword boundaries as a prior}
        \label{fig:subword-boundaries-small}
    \end{subfigure}
    \begin{subfigure}[t]{0.45\textwidth}
        \centering
        \includegraphics[width=\linewidth]{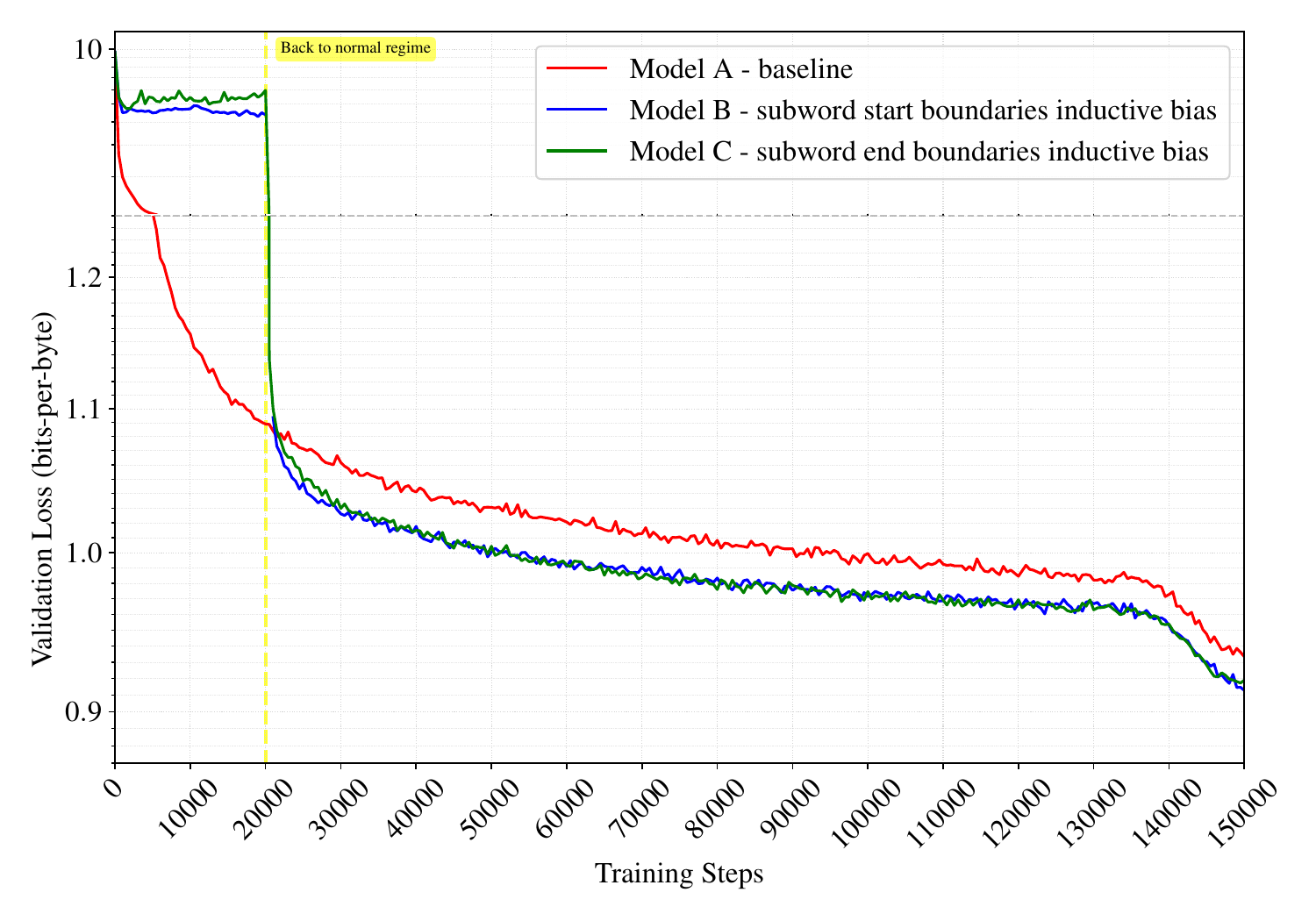}
        \caption{Subword boundaries as an inductive bias for $20$k steps}
        \label{fig:subword-boundaries-prior-small}
    \end{subfigure}
    \begin{subfigure}[t]{0.45\textwidth}
        \centering
        \includegraphics[width=\linewidth]{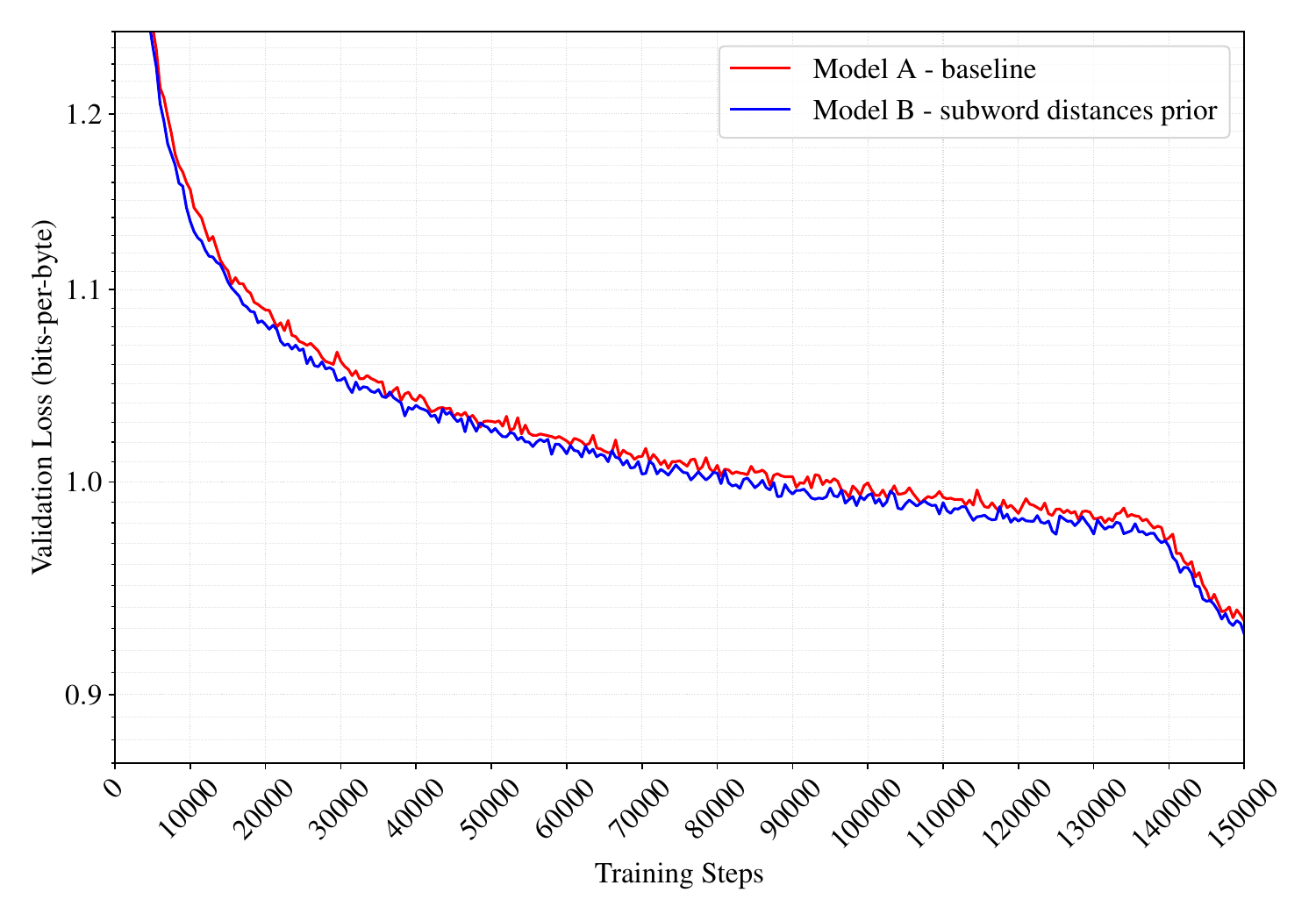}
        \caption{Subword distances as a prior}
        \label{fig:subword-distances-small}
    \end{subfigure}
    \begin{subfigure}[t]{0.45\textwidth}
        \centering
        \includegraphics[width=\linewidth]{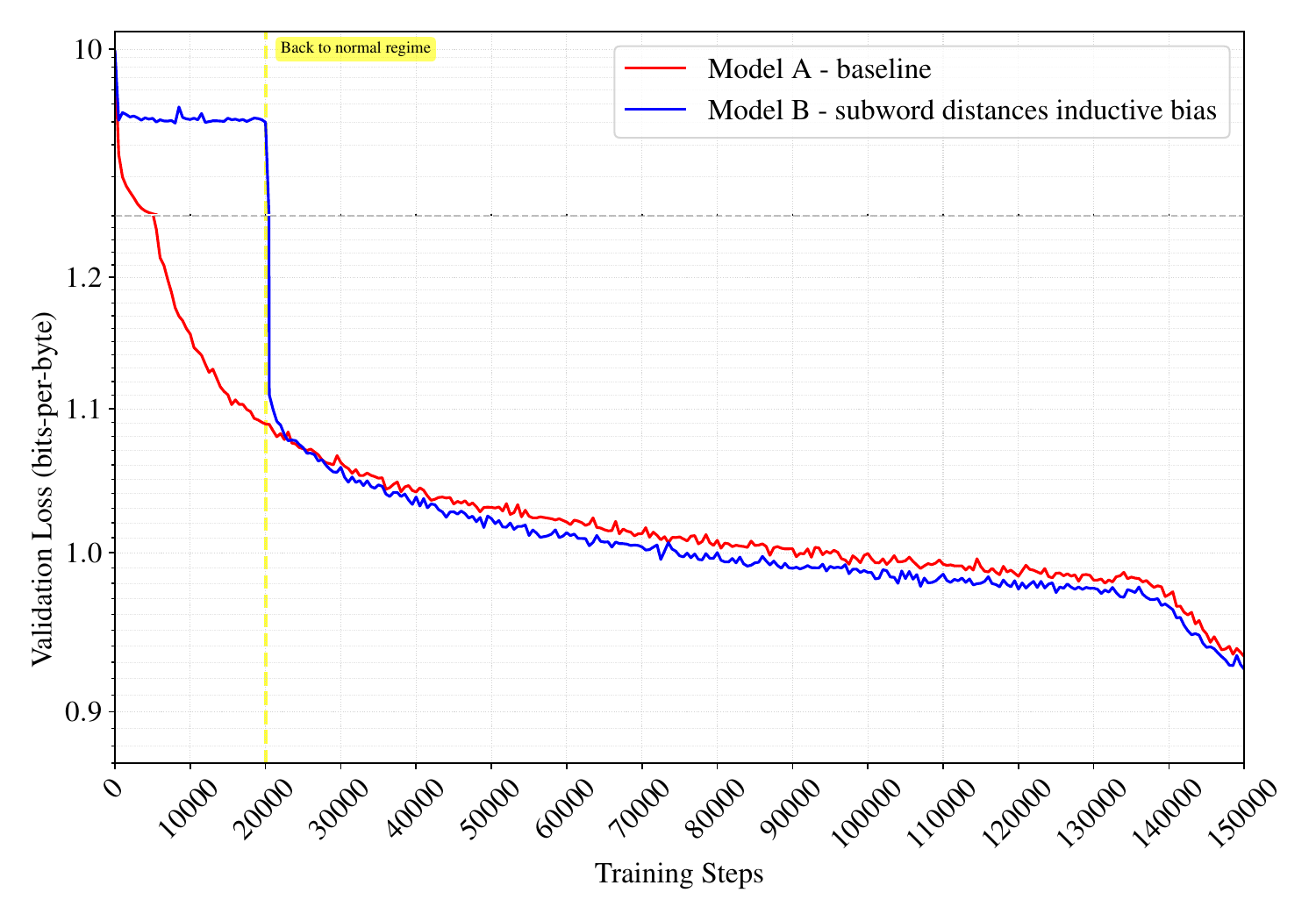}
        \caption{Subword distances as an inductive bias for $20$k steps for training only}
        \label{fig:subword-distances-prior-small}
    \end{subfigure}
    \begin{subfigure}[t]{0.45\textwidth}
        \centering
        \includegraphics[width=\linewidth]{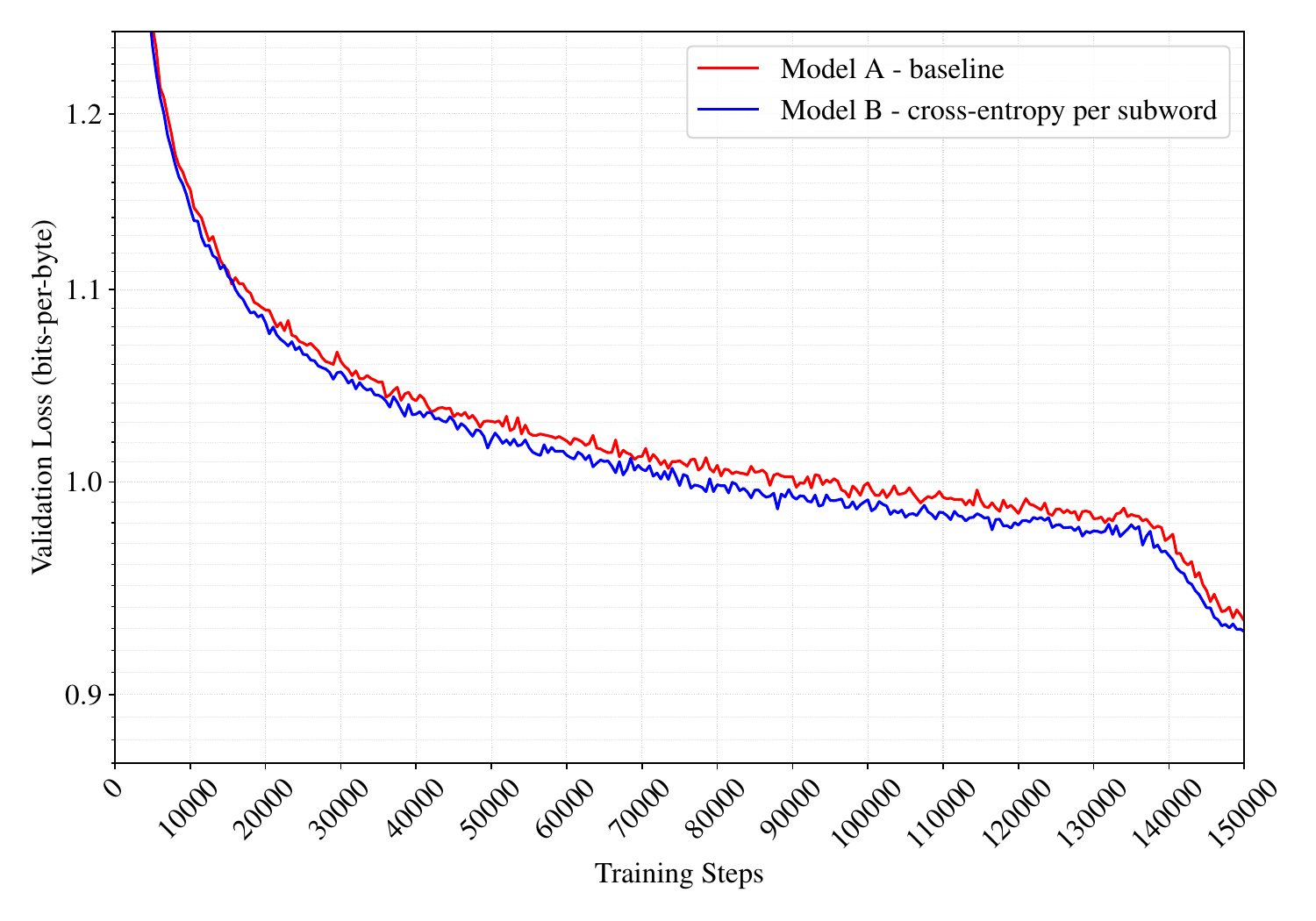}
    \caption{Optimizing for cross-entropy per subword}
    \label{fig:loss-per-subword-small}
    \end{subfigure}
    \begin{subfigure}[t]{0.45\textwidth}
        \centering
        \includegraphics[width=\linewidth]{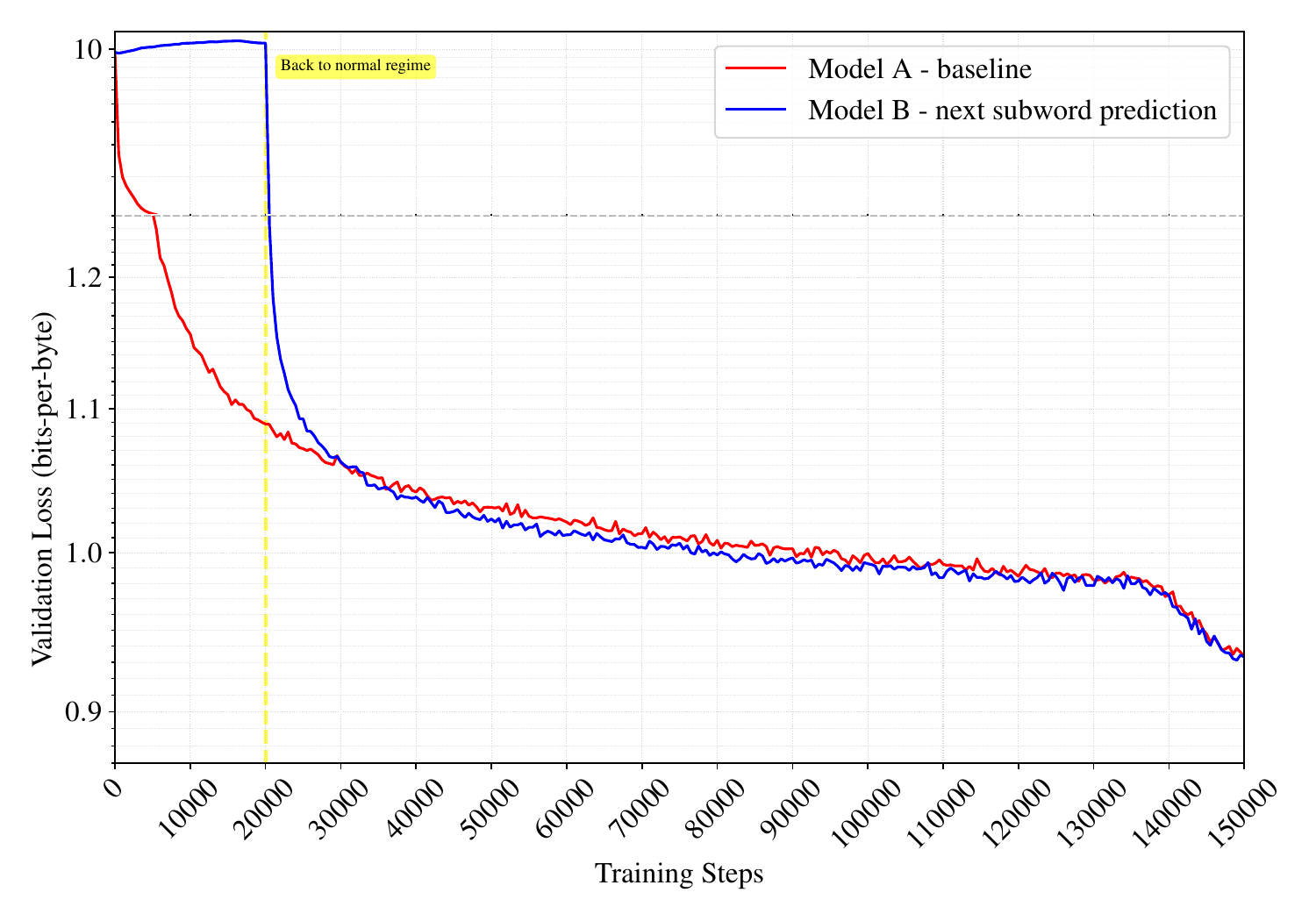}
    \caption{Optimizing for next subword prediction for $20$k steps}
    \label{fig:next-subword-prediction-small}
    \end{subfigure}

    \caption{Validation loss for different experiments with a $68$M parameters model}
    \label{fig:xps-small}
\end{figure*}

\end{document}